\documentclass[10pt,journal,twocolumn]{IEEEtran}

\usepackage{amsmath,amssymb}
\usepackage{array}
\usepackage{booktabs}
\usepackage{capt-of}
\usepackage{cite}
\usepackage{graphicx}
\usepackage{makecell}
\usepackage{multirow}
\usepackage[normalem]{ulem}
\usepackage{xcolor}
\usepackage{float}
\useunder{\uline}{\ul}{}

\definecolor{ieeeblue}{rgb}{0.21,0.49,0.74}
\usepackage[breaklinks,colorlinks,allcolors=ieeeblue]{hyperref}
\newcommand{\doi}[1]{\href{https://doi.org/#1}{\nolinkurl{#1}}}
\newcommand{\baselinecell}[1]{#1}

\newcommand{\stdcell}[2]{#1\,{\tiny$\pm$#2}}
\newcommand{\deltacell}[2]{#1\,{\tiny(#2)}}
\providecommand{\captionsetup}[2][]{}

\begin{document}

\title{LUMOS: Latent Universal Medical Priors for Segmentation}

\author{Zhuonan~Liang$^{1}$, Wei~Guo$^{1}$, Jie~Gan$^{1}$, Yaxuan~Song$^{1}$, Runnan~Chen$^{1}$, Hang~Chang$^{2,3}$, and Weidong~Cai$^{1}$%
\thanks{$^{1}$The University of Sydney, Sydney, NSW 2006, Australia.}%
\thanks{$^{2}$Biological System \& Engineering Division, Lawrence Berkeley National Laboratory, Berkeley, CA 94720, USA.}%
\thanks{$^{3}$Berkeley Biomedical Data Science Center, Lawrence Berkeley National Laboratory, Berkeley, CA 94720, USA.}%
\thanks{Corresponding author: Weidong Cai (tom.cai@sydney.edu.au).}}

\maketitle

\begin{abstract}
    General vision foundation models (VFMs) have been primarily developed on natural images, and their utility for medical image segmentation is therefore often considered to depend on costly adaptation or domain-specific fine-tuning. In this paper, we revisit this assumption from a different perspective: rather than requiring VFM segmentors to relearn visual regularities, we investigate whether the low-level visual priors necessary for anatomical delineation already lie dormant within general VFMs. We observe that frozen VFMs, despite lacking medical supervision, encode transferable visual regularities. These properties are not exclusive to natural images but are also fundamental to medical image understanding. Motivated by this observation, we propose Latent Universal Medical PriOrs for Segmentation (LUMOS), a novel framework that amplifies general VFM priors to conventional medical segmentors. LUMOS consists of two key components: (1) Pathfinder that distills visual cues from a frozen vision foundation model, and (2) Inspiror that sparks the conventional medical networks with spatial guidance from distilled visual regularities. In this way, the segmentor is relieved from learning complex visual regularities entirely from limited medical annotations and can instead focus on task-specific anatomical delineation. Across diverse medical datasets and token-based VFMs, LUMOS shows that general VFMs can serve as spatial prior generators when their frozen token spaces preserve patch-level pattern relevance. DINO provides stable matched-backbone gains, while SigLIP exposes VFM-specific sensitivity caused by its different token granularity and representation objective.
\end{abstract}


\section{Introduction}

Vision foundation models (VFMs) have reshaped computer vision through general representations learned at scale \cite{RN386,RN390,RN158,RN409}. Token-based VFMs, such as DINOv3 and SigLIP, provide dense features with global and local cues that may benefit medical segmentation \cite{RN390,RN386,RN385,RN406}. Yet their token semantics are not directly aligned with medical tasks \cite{RN395,RN385,RN409}, and fine-tuning them can require compute and annotations that remain scarce in medical imaging \cite{RN394}. Conventional medical segmentors, including convolutional and transformer variants, remain strong at modality-specific feature learning \cite{RN388,RN65,RN339,RN343,RN407,fan_structuring_2025}. The question is therefore how to use VFM information without forcing a medical segmentor to absorb mismatched semantics. DINO-style ViTs provide a clue: their self-attention exposes object layout and boundary information, and their patch features serve as dense descriptors for co-segmentation and correspondence \cite{RN412,RN413}. Recent medical studies likewise suggest that frozen DINO features preserve structural cues useful for lightweight segmentation readout \cite{RN409,RN414}. We therefore ask how the useful visual priors in native VFMs can be extracted and injected into medical segmentors in a controlled way.

\begin{figure*}[ht]
    \centering
    \includegraphics[width=0.9\textwidth]{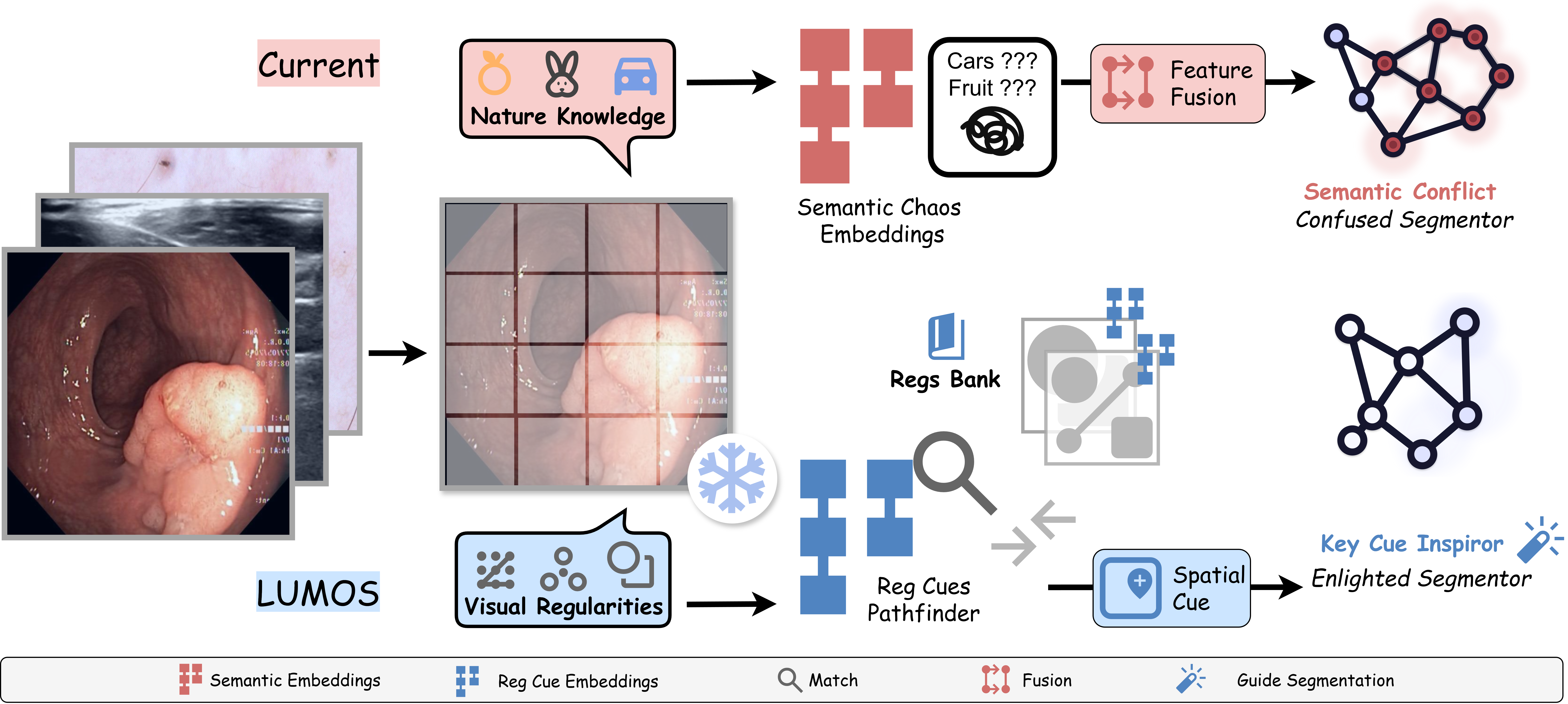}
    \caption{\textbf{Motivation of the proposed LUMOS framework.} We observe that frozen vision foundation models, despite lacking medical supervision, encode transferable visual regularities that are fundamental to medical image understanding. By leveraging these priors as spatial guidance, we can enhance the performance of conventional medical segmentation networks without requiring extensive fine-tuning of the foundation model.}
    \label{fig:teaser}

\end{figure*}

By leveraging the rich token representations from foundation models, we can efficiently guide the segmentation process while preserving the inductive biases of medical image segmentation architectures. The insights of this approach are twofold: (1) Foundation models can serve as visual-prior extractors that provide spatial guidance instead of direct medical semantics; (2) The dedicated medical network can focus on learning the semantic features relevant to medical images without being burdened by the need to adapt to the general representations. This raises a question: how can token features from general VFMs be converted into useful medical segmentation guidance without assuming that every VFM token is equally local or equally task aligned? To address this, we propose Latent Universal Medical PriOrs for Segmentation (LUMOS), a framework that leverages the general visual pattern representations of VFMs to guide conventional medical network training. Pathfinder extracts visual cue representations from a natively pretrained VFM and locates them on a spatial guide mask using a regularity-prototype Reg Book. Inspiror gates feature activations with the spatial mask, thereby injecting foundation-model priors while preserving the semantic biases and efficiency of medical image segmentation architectures. In parallel, the Reg Book adapts the frozen VFM token space into domain-specific regularity responses without fine-tuning the VFM encoder. Training combines standard segmentation losses with a guide supervision objective that aligns the guide mask to ground-truth regions, optionally augmented by a boundary-focused hinge loss to sharpen fine structures. Overall, our contributions are as follows:
\begin{enumerate}
    \item We provide a general VFM-as-guidance perspective for medical image segmentation, repositioning frozen VFMs as spatial pattern-prior generators rather than direct semantic segmentors.
    \item We propose the LUMOS framework with Pathfinder and Inspiror, which effectively leverages token features from VFM to create spatial masks that guide segmentation while maintaining the efficiency and inductive biases of medical image segmentation architectures.
    \item We demonstrate this perspective with DINOv3 and SigLIP across diverse medical datasets. DINO provides stable matched-backbone gains, while SigLIP's mixed behavior reveals how VFM-specific token granularity and patch-level relevance affect guide quality, supporting the pattern-prior interpretation.
\end{enumerate}

\section{Related Works}

Recent advances in medical image segmentation have increasingly explored the use of vision foundation models such as self-supervised DINO variants \cite{RN385,RN392,RN414,RN416} and the Segment Anything family \cite{RN393,RN400,RN401,RN402}, leveraging the rich semantic representations learned from large, diverse natural image corpora. In medical image analysis, these models hold promise for improving segmentation tasks, which are critical for diagnosis and treatment planning \cite{RN385,RN401,RN400,RN399}. However, the common practice of introducing foundation models as backbones for segmentation often overlooks the unique challenges of medical imaging, such as the need for precise boundary delineation, and significant modality shifts \cite{RN394,RN391,fan_revisiting_2025,zhang_rethinking_2023}. This can lead to suboptimal performance and inefficient training. Numerous works have evaluated the transferability and zero-shot capabilities of these foundation models in medical domains, highlighting both potential and challenges. Foundation models demonstrate significant generalization across modalities in zero-shot scenarios but often underperform without adaptation due to domain gap between natural and clinical images \cite{RN394,RN396,RN403,RN399}. Early adaptation strategies focused on fine-tuning VFMs or designing hybrid architectures that combine foundation backbones with task-specific decoders, showing improved performance at the cost of heavy supervision or prompt engineering \cite{RN402,RN393,RN401}. Parallel efforts on DINO-based adaptation have led to methods like MedDINOv3, SegDINO, Dino U-Net, and DINO-MVR, which explore multi-scale token aggregation, dense feature readout, or lightweight decoders together with frozen DINO backbones to balance efficiency and accuracy in segmentation tasks \cite{RN392,RN385,RN416,RN414}. Other research has examined the applicability of frozen foundation features for related dense prediction tasks and found that fixed representations can support training-free or test-time optimization strategies \cite{RN404}. Another relevant line of work uses spatial attention or gating to guide medical segmentation networks. Attention U-Net showed that attention gates can suppress irrelevant regions and highlight target structures with limited computational overhead \cite{RN415}, and later attention-based medical networks further exploit this idea within task-specific segmentation backbones \cite{RN407}. These methods support the value of spatial emphasis, but the guidance is learned inside the medical network itself. Despite the advances above, most foundation-model-based methods still treat the foundation model as a semantic predictor, decoder backbone, direct feature-fusion source, or fine-tuning target. LUMOS instead decouples roles: the frozen VFM provides visual-spatial emphasis through Reg Book responses, while the medical segmentation backbone remains responsible for semantic prediction, locality, multi-scale decoding, and boundary refinement.

\section{Methods}
\begin{figure*}[t]
    \centering
    \includegraphics[width=0.9\textwidth]{main.drawio.png}
    \caption{\textbf{Overview of the proposed LUMOS framework.} A frozen token-based VFM, instantiated by DINOv3 in the main experiments, serves as a guide generator by extracting dense token features that are converted into spatial guide masks via the Reg Book mechanism. These guide masks are then used to gate feature activations in the segmentation backbone, which produces the final segmentation output.}
    \label{fig:overview}

\end{figure*}

LUMOS is designed around the observation made in the Introduction: a frozen VFM may not be a reliable medical semantic segmentor, but its tokens can still preserve visual regularities that are useful for localization. Instead of fine-tuning the VFM or replacing the medical segmentation backbone, LUMOS separates the two roles. The VFM provides a soft spatial prior, while the medical segmentor remains responsible for anatomical classification, multi-scale decoding, and boundary refinement. This formulation is general to token-based VFMs, but the strength of the guide depends on how well each VFM preserves patch-level relevance to the target pattern. As shown in Figure~\ref{fig:overview}, LUMOS follows a simple pipeline: Pathfinder converts frozen VFM tokens into a guide mask through a Reg Book, and Inspiror injects this guide into a conventional segmentation backbone through soft feature gating.

\subsection{Overview and Problem Formulation}
Given a medical image $x\in\mathbb{R}^{B\times C_0\times H\times W}$, where $B$ is the batch size, $C_0$ is the number of input channels, and $H$ and $W$ are the image height and width, respectively, together with its annotation $y$, the goal is to predict a dense segmentation mask $\hat{y}$. LUMOS uses a frozen token-based VFM $F_{\phi}$ and a trainable medical segmentation network $S_{\theta}$. The VFM parameters $\phi$ are kept fixed so that the foundation model acts as a stable visual-prior extractor rather than a domain-adapted semantic predictor. The trainable parameters are the Reg Book and the segmentation backbone. This formulation lets LUMOS remain compatible with different token-based VFMs, such as DINOv3 and SigLIP, and with different medical segmentation backbones, such as UNet and nnWNet.

The full forward process has three steps. First, Pathfinder extracts patch-level tokens from $F_{\phi}$ and converts them into a spatial guide mask $G(x)$. Second, Inspiror resizes and injects $G(x)$ into the segmentation stream as a soft feature emphasis map. Third, the segmentation backbone predicts $\hat{y}$ from the guided features. The guide mask is supervised during training but is not used as a final prediction or a post-processing
constraint at inference time.

\subsection{Pathfinder: Distilling Cues}
Pathfinder addresses the semantic mismatch between natural-image VFMs and medical segmentation. Rather than asking the frozen VFM to directly classify anatomy, Pathfinder only asks it to indicate where visually coherent structures may lie. This is suitable for medical images because many segmentation targets are characterized by low-level cues such as local contrast, boundary continuity, texture change, and region coherence, which can be present in frozen VFM representations even without medical labels. The requirement is therefore not medical semantics in the VFM, but patch-level pattern relevance. A VFM with coarser tokenization or stronger global image-text alignment may provide less stable localization, yet its behavior still tests whether transferable pattern representations can be converted into spatial guidance.

For an input image $x$, the frozen VFM produces patch tokens:
\begin{equation}
    T = F_{\phi}(x), \quad T\in\mathbb{R}^{B\times L\times C}, \quad L=h\times w,
\end{equation}
where $h$ and $w$ denote the patch-grid resolution and $C$ is the token dimension. Pathfinder maintains a Reg Book with $K$ learnable prototypes $P\in\mathbb{R}^{K\times C}$. Each prototype represents a reusable visual-regularity pattern instead of an anatomical class. To stabilize this prototype memory during training, LUMOS applies exponential moving average (EMA) to the Reg Book:
\begin{equation}
    \bar{P}^{(t)} = \mu\bar{P}^{(t-1)} + (1-\mu)P^{(t)},
\end{equation}
where $P^{(t)}$ is the online Reg Book at iteration $t$, $\bar{P}^{(t)}$ is the EMA-smoothed Reg Book, and $\mu$ is the EMA momentum. The EMA-smoothed prototypes are used to produce stable regularity responses. Tokens and Reg Book prototypes are projected into a shared space, normalized, and compared by dot-product similarity:
\begin{equation}
    R_{i,k} = \operatorname{sim}(\hat{T}_{i}, \hat{\bar{P}}_{k}) = \hat{T}_{i}^{\top}\hat{\bar{P}}_{k},
\end{equation}
where $\hat{T}_{i}$ and $\hat{\bar{P}}_{k}$ denote normalized token and EMA-smoothed prototype embeddings. The prototype responses are then aggregated over the Reg Book and reshaped back to the VFM patch grid:
\begin{equation}
    G(x)=\operatorname{Norm}_{[0,1]}\left(
    \operatorname{Reshape}_{h,w}\left(\frac{1}{K}\sum_{k=1}^{K}R_{:,k}\right)
    \right),
\end{equation}
where $G(x)\in\mathbb{R}^{B\times1\times h\times w}$ is the guide mask. The normalization produces a continuous spatial emphasis map. High values indicate locations whose frozen tokens match the learned regularity prototypes, while low values indicate less relevant regions. In this way, Reg Book translates general VFM representations into task-aware but still soft spatial guidance.

\begin{table*}[!ht]
    \centering
    \caption{Baseline comparison across venues. LUMOS-W (DINO) achieves the best Dice on all datasets and the best IoU on Kvasir-SEG and TN3K.}
    \label{tab:baseline_results}
    \setlength{\tabcolsep}{2.5pt}
    \renewcommand{\arraystretch}{1.2}
    \footnotesize
    \begin{tabular}{@{}l@{\hspace{8pt}}c@{\hspace{10pt}}ccc@{\hspace{12pt}}ccc@{\hspace{12pt}}ccc@{}}
        \toprule
        \multirow{2}{*}{Method}
         & \multirow{2}{*}{Venue}
         & \multicolumn{3}{c}{Kvasir-SEG}
         & \multicolumn{3}{c}{ISIC 2017}
         & \multicolumn{3}{c}{TN3K}                                                              \\
        \cmidrule(lr){3-5}\cmidrule(lr){6-8}\cmidrule(lr){9-11}
         &                                & IoU$\uparrow$  & DSC$\uparrow$    & HD95$\downarrow$
         & IoU$\uparrow$                  & DSC$\uparrow$  & HD95$\downarrow$
         & IoU$\uparrow$                  & DSC$\uparrow$  & HD95$\downarrow$                    \\
        \midrule
        nnUNet \cite{RN343}
         & Nat. Methods'21
         & 83.92                          & 89.71          & 20.85
         & 82.30                          & 89.18          & 13.38
         & 72.81                          & 82.38          & 33.47                               \\
        SwinUNet \cite{RN342}
         & ECCVW'22
         & 40.75                          & 54.42          & 54.42
         & 81.24                          & 88.38          & 14.17
         & 52.28                          & 64.89          & 26.72                               \\
        H2Former \cite{RN340}
         & TMI'23
         & 83.87                          & 89.50          & 19.81
         & {\ul 82.56}                    & 89.36          & 13.04
         & 73.94                          & 82.67          & {\ul 21.91}                         \\
        U-KAN \cite{RN388}
         & AAAI'25
         & 63.81                          & 72.17          & 36.92
         & 74.62                          & 83.41          & 23.57
         & 69.60                          & 78.66          & 24.49                               \\
        nnWNet \cite{RN339}
         & CVPR'25
         & {\ul 84.58}                    & {\ul 89.98}    & {\ul 18.17}
         & \textbf{82.62}                 & {\ul 89.44}    & \textbf{12.73}
         & 66.46                          & 76.02          & 38.61                               \\
        SegDINO~\cite{RN385}
         & MICCAI'26
         & 80.64                          & 89.65          & 18.62
         & 77.60                          & 85.76          & 20.80
         & {\ul 74.43}                    & {\ul 83.18}    & \textbf{18.62}                      \\
        SAM2-UNet~\cite{RN411}
         & Vis. Intell.'26
         & 82.22                          & 89.24          & 19.97
         & 79.60                          & 86.40          & 16.91
         & 72.90                          & 81.70          & 23.60                               \\
        LUMOS-W (DINO)
         & Ours
         & \stdcell{\textbf{84.82}}{0.24}
         & \stdcell{\textbf{91.02}}{0.10}
         & \stdcell{\textbf{17.50}}{0.10}
         & \stdcell{82.32}{0.31}
         & \stdcell{\textbf{89.81}}{0.08}
         & \stdcell{{\ul 13.14}}{0.11}
         & \stdcell{\textbf{74.48}}{0.12}
         & \stdcell{\textbf{83.40}}{0.26}
         & \stdcell{24.08}{0.17} \\
        \bottomrule
    \end{tabular}

\end{table*}

\subsection{Inspiror: Injecting Guidance}
Inspiror injects the Pathfinder guide into a medical segmentation backbone through soft gating rather than feature fusion. This distinction is important because frozen VFM features carry natural-image semantic biases that may conflict with the semantic space learned by a medical segmentor. Instead of concatenating or fusing these features with backbone activations, Inspiror uses only the guide mask as a spatial modulation signal, preserving the segmentor's responsibility for anatomical prediction. Let $X_{\ell}$ be the feature map at an input or intermediate layer $\ell$ of the segmentation backbone. The guide mask is resized to the feature resolution and broadcast along the channel dimension:
\begin{equation}
    \tilde{G}_{\ell} = \operatorname{Up}_{\ell}(G(x)).
\end{equation}
Inspiror then applies soft multiplicative gating:
\begin{equation}
    X_{\ell}^{\prime}=X_{\ell}\odot \tilde{G}_{\ell},
\end{equation}
where $\odot$ denotes element-wise multiplication. This operation encourages the segmentor to allocate more representational capacity to visually relevant regions while preserving the backbone's own semantic learning process. Because the guide is continuous and learned jointly with the segmentation objective, it acts as a soft emphasis field rather than a semantic label or hard constraint; the backbone can still recover from imperfect guide responses and refine structures using its local and multi-scale inductive biases.

This design differs from directly concatenating foundation-model features or replacing the medical backbone with a VFM decoder. Direct feature fusion forces the segmentor to absorb semantic representations whose categories and statistics may not match medical images, potentially introducing semantic conflict. In contrast, Inspiror transfers only spatial emphasis. The segmentation network therefore remains lightweight and task-specific, while Pathfinder supplies a VFM-derived prior about where the network should attend.

\subsection{Optimization Objectives}
LUMOS is optimized with a segmentation objective and a guide-alignment objective. The segmentation loss supervises the final prediction $\hat{y}$ from the medical backbone:
\begin{equation}
    \mathcal{L}_{\mathrm{seg}} = \mathcal{L}_{\mathrm{seg}}(\hat{y}, y),
\end{equation}
where the concrete form follows the backbone training protocol, such as Dice loss or cross-entropy loss for dense prediction. To make the guide mask informative for the target task, the annotation is resized to the guide resolution and used to supervise Pathfinder:
\begin{equation}
    \mathcal{L}_{\mathrm{guide}}
    =-\frac{1}{|\Omega|}\sum_{i\in\Omega}
    \left[y^{G}_{i}\log g_i + (1-y^{G}_{i})\log(1-g_i)\right],
\end{equation}
where $g_i$ is the guide-mask value at patch location $i$, $y^{G}_{i}$ is the corresponding resized ground-truth label, and $\Omega$ denotes all guide-grid locations. The full objective is
\begin{equation}
    \mathcal{L}
    =\mathcal{L}_{\mathrm{seg}}
    +\lambda_{\mathrm{guide}}\mathcal{L}_{\mathrm{guide}}
    +\lambda_{\mathrm{bd}}\mathcal{L}_{\mathrm{bd}},
\end{equation}
where $\lambda_{\mathrm{guide}}$ balances guide supervision and $\mathcal{L}_{\mathrm{bd}}$ denotes an optional boundary-focused regularizer. When enabled, the boundary term encourages sharper transitions around annotated contours; otherwise $\lambda_{\mathrm{bd}}$ is set to zero. This composite training objective aligns the VFM-derived guide with target regions while keeping the final segmentation decision in the conventional medical backbone.

\subsubsection{VFM-Specific Pattern Enhancement}
The default LUMOS setting keeps the VFM encoder frozen and trains only the Reg Book and segmentation backbone. We additionally consider an optional VFM-specific pattern enhancement (SPE) to examine whether a small trainable residual path can make the frozen token space more responsive to dataset-specific visual regularities. Instead of viewing this variant as direct VFM fine-tuning, we formulate it by its backpropagation path. Let $\psi$ denote the lightweight enhancement parameters and $T_{\psi}$ denote the enhanced VFM tokens. The original VFM weights remain fixed, and only $\psi$ is updated:
\begin{equation}
    \psi^{(t+1)}
    =
    \psi^{(t)}
    -\eta
    \frac{\partial \mathcal{L}}{\partial \psi},
    \quad
    \frac{\partial \mathcal{L}}{\partial \psi}
    =
    \frac{\partial \mathcal{L}}{\partial G}
    \frac{\partial G}{\partial R}
    \frac{\partial R}{\partial T_{\psi}}
    \frac{\partial T_{\psi}}{\partial \psi}.
\end{equation}
Here $R_{i,k}=\hat{T}_{\psi,i}^{\top}\hat{\bar{P}}_k$ is the Reg Book response between an enhanced token and the EMA-smoothed regularity prototype. The terms $\partial G/\partial R$ and $\partial R/\partial T_{\psi}$ show that the backpropagated signal is filtered by Reg Book matching before it reaches the enhancement parameters. Thus, the update is driven by prototype-aligned spatial regularities rather than by unconstrained semantic feature fusion. This filtering helps alleviate the semantic conflict discussed above: the VFM is not asked to inject its full natural-image semantic representation into the medical segmentor, but only to adjust token directions that contribute to Reg Book-guided spatial emphasis. This variant therefore adjusts the pattern responses used by Pathfinder without changing the basic LUMOS pipeline.

Because the enhancement slightly shifts the VFM token space, its benefit is not guaranteed. It may improve guidance when the target dataset contains consistent visual patterns that can be captured by a compact residual update, but it can also disturb useful frozen priors or overfit to dataset-specific appearance. We therefore treat this enhancement as an auxiliary analysis rather than the main evidence for LUMOS; matched-backbone results with frozen VFMs remain the primary validation of the guide mechanism.

\section{Experiments and Results}

\subsection{Datasets and Evaluation Metrics}
The experiments are conducted on several publicly available medical imaging datasets, including Kvasir \cite{RN338}, ISIC \cite{RN336}, and TN3K \cite{RN337}. \textbf{Kvasir-SEG} consists of 1,000 colonoscopy images with corresponding pixel-level binary masks for polyp segmentation, and we use an 800/200 train/test split. Image resolutions vary across samples. \textbf{ISIC 2017} contains dermoscopic images with expert-annotated lesion masks, and we follow the original 2001/151 split. \textbf{TN3K} includes 3,493 thyroid ultrasound images from 2,421 patients with pixel-level nodule annotations, and we follow the original 2879/614 split. These datasets cover a range of medical imaging modalities and segmentation tasks, providing a comprehensive evaluation of the proposed LUMOS framework. We use standard evaluation metrics for medical image segmentation, including Dice coefficient, Intersection over Union (IoU), and HD95 Score, to assess the performance of our model.

\subsection{Implementation Details}
We implement the LUMOS framework using PyTorch. The DINOv3 backbone is initialized with pretrained weights from the official huggingface repository\footnote{https://huggingface.co/facebook/dinov3-vits16-pretrain-lvd1689m}. And the weights of SigLIP2 sources from huggingface \footnote{https://huggingface.co/google/siglip2-base-patch32-256}. VFM backbone is kept frozen during training, while the Reg Book and segmentation backbone are trained from scratch. EMA is applied to the Reg Book prototypes during training to stabilize guide generation. The input images are resized to a fixed resolution of 256x256 for training. We use an initial learning rate of 1e-4 and a batch size of 4. For the optional VFM-specific pattern enhancement variant, AdamW is applied to the enhancement parameters with a 5e-4 learning rate. The model is trained for 400 epochs, and the best model is selected based on validation performance. Augmentation and inference follow the nnUNet \cite{RN343} implementation. More details can be found on our project page\footnote{https://anonymous.4open.science/r/MedToken-FC36/}. The experiments are conducted on a single NVIDIA RTX 6000 Ada GPU with 40GB of memory. All results are averaged over three runs with different random seeds to ensure robustness and reproducibility.

\subsection{Comparison with Existing Methods}

Table~\ref{tab:baseline_results} compares LUMOS-W with recent medical segmentation networks and foundation-model-related methods under the same evaluation protocol. The comparison is intended to test the first contribution: whether a frozen VFM can be useful when used as a spatial-prior generator rather than as a direct semantic segmentor. LUMOS-W achieves the best Dice on Kvasir-SEG, ISIC 2017, and TN3K, and obtains the best IoU on Kvasir-SEG and TN3K. Compared with its nnWNet backbone baseline shown in the table, LUMOS-W improves Kvasir-SEG by $+0.24$ IoU and $+1.04$ Dice, and improves TN3K by $+8.02$ IoU and $+7.38$ Dice. These gains indicate that the VFM-derived guide can add complementary localization information even when the final prediction is still produced by a medical segmentation backbone.

The baseline comparison also shows where the claim should be bounded. LUMOS-W is not the best method for every boundary-distance entry: nnWNet remains slightly better on ISIC 2017 HD95, and SegDINO gives the lowest TN3K HD95. This behavior is reasonable because LUMOS supplies a low-resolution soft guide rather than a boundary-specialized decoder or post-processing module. Dice and IoU mainly reflect region-level overlap, whereas HD95 is sensitive to a small number of contour outliers. Thus, the suboptimal HD95 results do not contradict the proposed VFM-as-guidance formulation; instead, they show that a spatial prior improves region localization most reliably, while final boundary accuracy still depends on the backbone, decoder, and dataset-specific contour ambiguity. For this reason, Table~\ref{tab:baseline_results} establishes external competitiveness, while the matched-backbone ablations below provide the more direct test of whether the VFM-derived guide contributes beyond the segmentor itself.

\begin{table*}[!ht]
    \centering
    \caption{Matched-backbone ablation. Deltas relative to each UNet or nnWNet baseline show that DINO-based LUMOS consistently improves Dice, with larger gains in weaker settings.}
    \label{tab:ablation_guidino_backbones}
    \setlength{\tabcolsep}{4.7pt}
    \renewcommand{\arraystretch}{1.24}
    \footnotesize
    \begin{tabular}{@{}llccc@{\hspace{5pt}}ccc@{\hspace{5pt}}ccc@{}}
        \toprule
        \multirow{2}{*}{Backbone}
         & \multirow{2}{*}{Method}
         & \multicolumn{3}{c}{Kvasir-SEG}
         & \multicolumn{3}{c}{ISIC 2017}
         & \multicolumn{3}{c}{TN3K}                                                             \\
        \cmidrule(lr){3-5}\cmidrule(lr){6-8}\cmidrule(lr){9-11}
         &                                & IoU$\uparrow$ & DSC$\uparrow$    & HD95$\downarrow$
         & IoU$\uparrow$                  & DSC$\uparrow$ & HD95$\downarrow$
         & IoU$\uparrow$                  & DSC$\uparrow$ & HD95$\downarrow$                    \\
        \midrule
        \multirow[c]{3}{*}[-0.45ex]{UNet}
         & Baseline \cite{RN65}
         & \baselinecell{70.29}
         & \baselinecell{79.16}
         & \baselinecell{41.58}
         & \baselinecell{72.95}
         & \baselinecell{81.87}
         & \baselinecell{41.58}
         & \baselinecell{70.65}
         & \baselinecell{79.45}
         & \baselinecell{24.59}                                                                 \\
         & LUMOS (SigLIP)
         & \deltacell{78.69}{+8.40}
         & \deltacell{86.79}{+7.63}
         & \deltacell{26.33}{-15.25}
         & \deltacell{76.82}{+3.87}
         & \deltacell{86.71}{+4.84}
         & \deltacell{16.59}{-24.99}
         & \deltacell{69.63}{-1.02}
         & \deltacell{79.50}{+0.05}
         & \deltacell{28.13}{+3.54} \\
         & LUMOS (DINO)
         & \deltacell{83.18}{+12.89}
         & \deltacell{89.51}{+10.35}
         & \deltacell{17.78}{-23.80}
         & \deltacell{81.24}{+8.29}
         & \deltacell{88.78}{+6.91}
         & \deltacell{17.86}{-23.72}
         & \deltacell{72.09}{+1.44}
         & \deltacell{81.41}{+1.96}
         & \deltacell{24.97}{+0.38} \\
        \midrule
        \multirow[c]{3}{*}[-0.45ex]{nnWNet}
         & Baseline \cite{RN339}
         & \baselinecell{84.58}
         & \baselinecell{89.98}
         & \baselinecell{18.17}
         & \baselinecell{82.62}
         & \baselinecell{89.44}
         & \baselinecell{12.73}
         & \baselinecell{66.46}
         & \baselinecell{76.02}
         & \baselinecell{38.61}                                                                 \\
         & LUMOS (SigLIP)
         & \deltacell{82.35}{-2.23}
         & \deltacell{89.31}{-0.67}
         & \deltacell{21.72}{+3.55}
         & \deltacell{81.01}{-1.61}
         & \deltacell{88.76}{-0.68}
         & \deltacell{14.40}{+1.67}
         & \deltacell{72.52}{+6.06}
         & \deltacell{82.57}{+6.55}
         & \deltacell{24.30}{-14.31} \\
         & LUMOS (DINO)
         & \deltacell{84.82}{+0.24}
         & \deltacell{91.02}{+1.04}
         & \deltacell{17.50}{-0.67}
         & \deltacell{82.32}{-0.30}
         & \deltacell{89.81}{+0.37}
         & \deltacell{13.14}{+0.41}
         & \deltacell{74.48}{+8.02}
         & \deltacell{83.40}{+7.38}
         & \deltacell{24.08}{-14.53} \\
        \bottomrule
    \end{tabular}
\end{table*}

\subsection{Ablation Studies}

In ablation studies, we first use matched-backbone comparisons to isolate whether the guide itself contributes beyond the underlying segmentor. We then test the key design choice in Inspiror, namely soft gating instead of direct feature fusion, and study the spatial resolution of the guide mask. Finally, we analyze optional VFM-specific pattern enhancement and a reduced-annotation setting to clarify when the guidance is most useful and when it becomes less stable.

\begin{table*}[ht]
    \centering
    \caption{Optional VFM-specific pattern enhancement for DINO-derived guidance. The enhancement improves several metrics but remains dataset-dependent; matched frozen-VFM LUMOS-W remains the main guide ablation.}
    \label{tab:specific_pattern_enhancement}
    \setlength{\tabcolsep}{7.9pt}
    \setlength{\extrarowheight}{1pt}
    \renewcommand{\arraystretch}{1.04}
    \footnotesize
    \begin{tabular}{@{}lccc@{\hspace{5pt}}ccc@{\hspace{5pt}}ccc@{}}
        \toprule
        \multirow{2}{*}{Method}
         & \multicolumn{3}{c}{Kvasir-SEG}
         & \multicolumn{3}{c}{ISIC 2017}
         & \multicolumn{3}{c}{TN3K}                                          \\
        \cmidrule(lr){2-4}\cmidrule(lr){5-7}\cmidrule(lr){8-10}
         & IoU$\uparrow$                  & DSC$\uparrow$ & HD95$\downarrow$
         & IoU$\uparrow$                  & DSC$\uparrow$ & HD95$\downarrow$
         & IoU$\uparrow$                  & DSC$\uparrow$ & HD95$\downarrow$ \\
        \midrule
        Seg-DINO \cite{RN385}
         & 80.64                          & 89.65         & 18.62
         & 77.60                          & 85.76         & 20.80
         & 74.43                          & 83.18         & 18.62            \\

        Seg-LUMOS
         & \deltacell{79.10}{-1.54}
         & \deltacell{87.23}{-2.42}
         & \deltacell{18.64}{+0.02}
         & \deltacell{81.08}{+3.48}
         & \deltacell{89.23}{+3.47}
         & \deltacell{16.42}{-4.38}
         & \deltacell{61.67}{-12.76}
         & \deltacell{73.08}{-10.10}
         & \deltacell{16.13}{-2.51}
        \\
        Seg-LUMOS (SPE)
         & \deltacell{87.37}{+6.73}
         & \deltacell{92.72}{+3.07}
         & \deltacell{12.16}{-6.46}
         & \deltacell{79.54}{+1.94}
         & \deltacell{88.11}{+2.35}
         & \deltacell{16.56}{-4.24}
         & \deltacell{74.58}{+0.15}
         & \deltacell{83.56}{+0.38}
         & \deltacell{21.41}{+2.79}
        \\
        \midrule
        nnWNet \cite{RN339}
         & 84.58                          & 89.98         & 18.17
         & 82.62                          & 89.44         & 12.73
         & 66.46                          & 76.02         & 38.61            \\
        LUMOS-W
         & \deltacell{84.82}{+0.24}
         & \deltacell{90.86}{+0.88}
         & \deltacell{17.50}{-0.67}
         & \deltacell{82.32}{-0.30}
         & \deltacell{89.81}{+0.37}
         & \deltacell{13.14}{+0.41}
         & \deltacell{74.48}{+8.03}
         & \deltacell{83.40}{+7.38}
         & \deltacell{24.08}{-14.53}
        \\
        LUMOS-W (SPE)
         & \deltacell{86.49}{+1.91}
         & \deltacell{92.09}{+2.11}
         & \deltacell{12.40}{-5.77}
         & \deltacell{82.65}{+0.03}
         & \deltacell{89.83}{+0.40}
         & \deltacell{13.42}{+0.28}
         & \deltacell{73.04}{+6.58}
         & \deltacell{82.53}{+6.51}
         & \deltacell{31.20}{-7.41}
        \\
        \bottomrule
    \end{tabular}

\end{table*}
\subsubsection{Backbone-Level Guidance}

Table~\ref{tab:ablation_guidino_backbones} provides the main causal evidence for LUMOS because each guided model is compared with its own backbone under the same protocol. With DINO guidance, LUMOS improves Dice in all six matched settings. The largest improvements occur when the unguided backbone has clear room for localization improvement, such as UNet on Kvasir-SEG and ISIC 2017 or nnWNet on TN3K. This pattern supports the second contribution: Pathfinder and Inspiror do not replace the medical backbone, but provide a complementary spatial prior that helps the backbone allocate capacity to target regions.

The same table also explains several suboptimal results. As with nnWNet on ISIC 2017, DINO guidance yields only a small Dice gain and slightly worse IoU and HD95. This is an expected saturation effect: the soft guide can improve foreground coverage, but it cannot guarantee that every boundary point improves, especially when the backbone already has strong inductive bias for that dataset. The UNet+LUMOS result on TN3K shows a similar tradeoff, improving Dice and IoU while slightly increasing HD95. These cases indicate that LUMOS is most reliable as a region-localization aid, whereas HD95 can still be affected by isolated boundary errors.

SigLIP further demonstrates that the benefit depends on VFM token quality. SigLIP guidance improves UNet on Kvasir-SEG and ISIC 2017 and substantially improves nnWNet on TN3K, but it reduces several nnWNet metrics on Kvasir-SEG and ISIC 2017. This mixed behavior is consistent with the third contribution: general VFMs can provide spatial priors, but their usefulness depends on how well their token spaces preserve patch-level relevance for the target morphology. The SigLIP2 Patch32 configuration and image-text alignment objective emphasize different visual statistics from DINO, which can make local medical boundaries less stable while still preserving useful pattern responses for some datasets. Therefore, the ablation supports a precise interpretation rather than an unconditional universality claim: LUMOS is effective when the frozen VFM token space contains local visual regularities that can be converted into reliable spatial guidance.

\begin{figure*}[!ht]
    \centering
    \includegraphics[width=\textwidth]{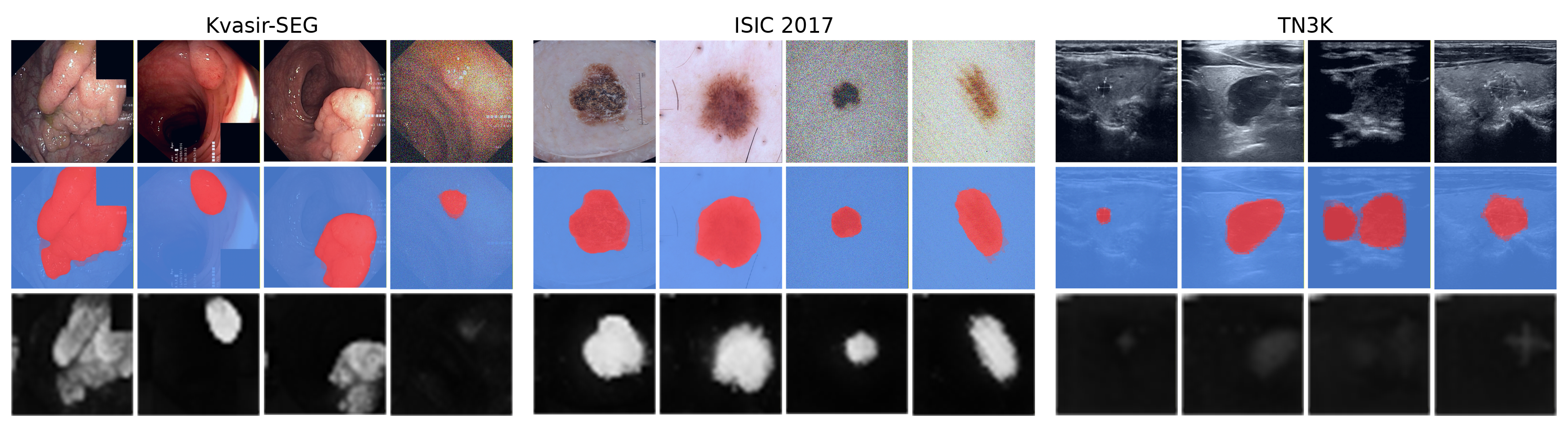}
    \caption{\textbf{Qualitative results for segmentation.} \textbf{First row}: original input image. \textbf{Second row}: actual segmentation masks. \textbf{Third row}: guidance mask from LUMOS. The guidance mask highlights the rough location of the target, which can assist the segmentation backbone in focusing on relevant regions.}
    \label{fig:vis}
\end{figure*}
\subsubsection{Guide Injection and Mask Resolution}

Table~\ref{tab:injection_ablation} tests whether the guide should be injected as soft gating or by direct fusion. Direct fusion degrades all three Kvasir-SEG metrics relative to the nnWNet baseline, while soft gating improves IoU, Dice, and HD95. This result supports the design motivation of Inspiror: the VFM should not force its full natural-image feature semantics into the medical segmentation stream. Instead, the guide is more effective when it acts as a continuous spatial emphasis map and leaves anatomical classification to the medical backbone.

\begin{center}
    \begin{minipage}{\linewidth}
        \centering
        \captionsetup{hypcap=false}
        \captionof{table}{Guide-injection design on Kvasir-SEG. Soft gating improves Dice and HD95 over direct fusion and the nnWNet-based baseline.}
        \label{tab:injection_ablation}
        \setlength{\tabcolsep}{7pt}
        \renewcommand{\arraystretch}{1.04}
        \footnotesize
        \begin{tabular}{@{}lccc@{}}
            \toprule
            Method       & IoU$\uparrow$ & DSC$\uparrow$ & HD95$\downarrow$ \\
            \midrule
            Baseline-W   & 84.58         & 89.98         & 18.17            \\
            Direct fused & \deltacell{83.82}{-0.76} & \deltacell{88.44}{-1.54} & \deltacell{22.01}{+3.84}          \\
            Soft gating  & \deltacell{\textbf{84.82}}{+0.24} & \deltacell{\textbf{91.02}}{+1.04} & \deltacell{\textbf{17.50}}{-0.67} \\
            \bottomrule
        \end{tabular}
    \end{minipage}
\end{center}
Table~\ref{tab:guide_mask_size_ablation} shows that guide resolution is also important. In the compact main-text sweep, the $22 \times 22$ guide gives the best TN3K IoU and Dice while maintaining comparable HD95 to nearby resolutions. Coarser guides can miss small or irregular target regions, whereas a larger $32 \times 32$ guide does not further improve performance and may introduce noisier local emphasis after resizing. This confirms that the guide mask is not intended to be a high-resolution segmentation prediction; it should provide enough spatial structure to guide the backbone without over-constraining the final decoder. The injection and resolution ablations align with the method design that Pathfinder extracts regularity responses from frozen tokens, and Inspiror transfers them as spatial emphasis without changing the semantic role of the medical backbone.

\begin{center}
    \begin{minipage}{\linewidth}
        \centering
        \captionsetup{hypcap=false}
        \captionof{table}{Guide-mask resolution on TN3K in the compact main-text sweep. The $22 \times 22$ mask gives the best final IoU and DSC among these reported settings and is selected as the overlap-oriented tradeoff.}
        \label{tab:guide_mask_size_ablation}
        \setlength{\tabcolsep}{14.0pt}
        \renewcommand{\arraystretch}{1.06}
        \footnotesize
        \begin{tabular}{@{}cccc@{}}
            \toprule
            \makecell{Guide Mask                                                                 \\Size}
                           & \makecell{IoU$\uparrow$}
                           & DSC$\uparrow$
                           & HD95$\downarrow$                                                    \\
            \midrule
            $8 \times 8$   & 73.34          & 82.69          & 24.65          \\
            $11 \times 11$ & 72.80          & 81.94          & 24.76          \\

            $22 \times 22$ & \textbf{74.48} & \textbf{83.40} & \textbf{24.08} \\
            $32 \times 32$ & 72.81          & 82.19          & 25.15          \\
            \bottomrule
        \end{tabular}
    \end{minipage}
\end{center}

\subsubsection{VFM-Specific Pattern Enhancement}

\begin{table*}[ht]
    \centering
    \caption{Reduced-annotation setting with 30\% training masks. LUMOS-W improves nnWNet across datasets, with especially large Kvasir-SEG gains and lower HD95 on all three benchmarks.}
    \label{tab:reduced_annotation}
    \setlength{\tabcolsep}{5.5pt}
    \renewcommand{\arraystretch}{1.12}
    \begin{tabular}{@{}l@{\hspace{0.8em}}ccc@{\hspace{1.2em}}ccc@{\hspace{1.2em}}ccc@{}}
        \toprule
                                & \multicolumn{3}{c}{Kvasir-SEG}
                                & \multicolumn{3}{c}{ISIC 2017}
                                & \multicolumn{3}{c}{TN3K}                                                              \\
        \cmidrule(lr){2-4}\cmidrule(lr){5-7}\cmidrule(l){8-10}
        Method                  & IoU$\uparrow$            & DSC$\uparrow$            & HD95$\downarrow$
                                & IoU$\uparrow$            & DSC$\uparrow$            & HD95$\downarrow$
                                & IoU$\uparrow$            & DSC$\uparrow$            & HD95$\downarrow$          \\
        \midrule
        nnWNet                  & 72.56                    & 81.59                    & 37.34
                                & 72.20                    & 80.93                    & 41.02
                                & 66.41                    & 75.33                    & 56.27                      \\
        LUMOS-W (DINO)          & \deltacell{80.50}{+7.94} & \deltacell{87.34}{+5.75} & \deltacell{19.91}{-17.43}
                                & \deltacell{75.09}{+2.89} & \deltacell{85.02}{+4.09} & \deltacell{28.92}{-12.10}
                                & \deltacell{67.43}{+1.02} & \deltacell{78.20}{+2.87} & \deltacell{33.67}{-22.60} \\
        \bottomrule
    \end{tabular}
\end{table*}

Table~\ref{tab:specific_pattern_enhancement} studies whether lightly adapting the VFM token response improves the guide. In the Seg-DINO branch, Seg-LUMOS without SPE improves ISIC 2017 and lowers TN3K HD95, but it decreases Kvasir-SEG and TN3K overlap metrics. After adding SPE, Kvasir-SEG improves substantially and ISIC 2017 remains better than Seg-DINO, but TN3K shows a tradeoff: overlap becomes slightly higher than Seg-DINO, while HD95 becomes worse. In the nnWNet branch, SPE strengthens Kvasir-SEG, is nearly neutral on ISIC 2017, and remains beneficial over nnWNet on TN3K but weaker than frozen LUMOS-W for TN3K overlap and HD95. These mixed results clarify why SPE is treated as an auxiliary option rather than the default evidence for LUMOS. A small enhancement path can sharpen pattern responses when the dataset contains consistent visual regularities, as in Kvasir-SEG. However, it can also disturb useful frozen priors, especially when boundaries are weak, noisy, or highly variable. The frozen-VFM backbone ablation is therefore the cleaner demonstration that useful VFM priors can be extracted without requiring the VFM itself to become a medical semantic segmentor.

\subsubsection{Reduced-Annotation Setting}

Finally, Table~\ref{tab:reduced_annotation} evaluates the 30\% annotation setting. This setting supports that when dense masks are limited, the backbone benefits from an external spatial prior that reduces the burden of learning localization entirely from annotated medical data. Across the ablations, the benefit is strongest when localization is difficult or supervision is scarce, and weaker when final performance depends on fine contour modeling beyond the resolution and reliability of the guide mask.

\subsection{Qualitative Analysis}

Figure~\ref{fig:vis} provides qualitative evidence for the mechanism behind the quantitative results. The DINO-derived guide generally highlights the target region rather than producing an exact segmentation mask. This is the desired behavior: the guide supplies coarse spatial emphasis, and the medical segmentation backbone remains responsible for semantic prediction and boundary refinement. Cases where the guide is broader or less boundary-aligned help explain the HD95 tradeoffs observed above, because a low-resolution guide can improve region coverage without perfectly matching the final contour.

\section{Conclusion}
LUMOS demonstrates that general vision foundation models can benefit medical segmentation without being converted into segmentors. By extracting regularity responses from frozen VFM tokens through Pathfinder and injecting them as soft spatial gates through Inspiror, LUMOS supplies complementary localization priors while leaving anatomical prediction to the medical backbone. The experiments support this claim with competitive baseline results, and strong improvements under reduced annotation. The mixed ablation results demonstrate LUMOS as a region-localization aid, not a guaranteed boundary-refinement module, and its strength depends on token locality, guide quality, and dataset morphology. These findings make frozen VFM priors a practical guidance source for medical segmentation, with finer boundary modeling remaining the main direction for extension.

\pagebreak
{\small
    \bibliographystyle{IEEEtran}
    \bibliography{ref}
}
\clearpage
\appendices
\providecommand{\MainReducedAnnotationTable}{Table~\ref{tab:reduced_annotation}}

\twocolumn[{
\section{Low-Annotation Curve Analysis}
\label{sec:appendix_low_annotation}

\noindent Figure~\ref{fig:low_annotation_curves} reports the full training dynamics for the reduced-annotation setting summarized in \MainReducedAnnotationTable.
\begin{center}
    \includegraphics[width=0.98\textwidth]{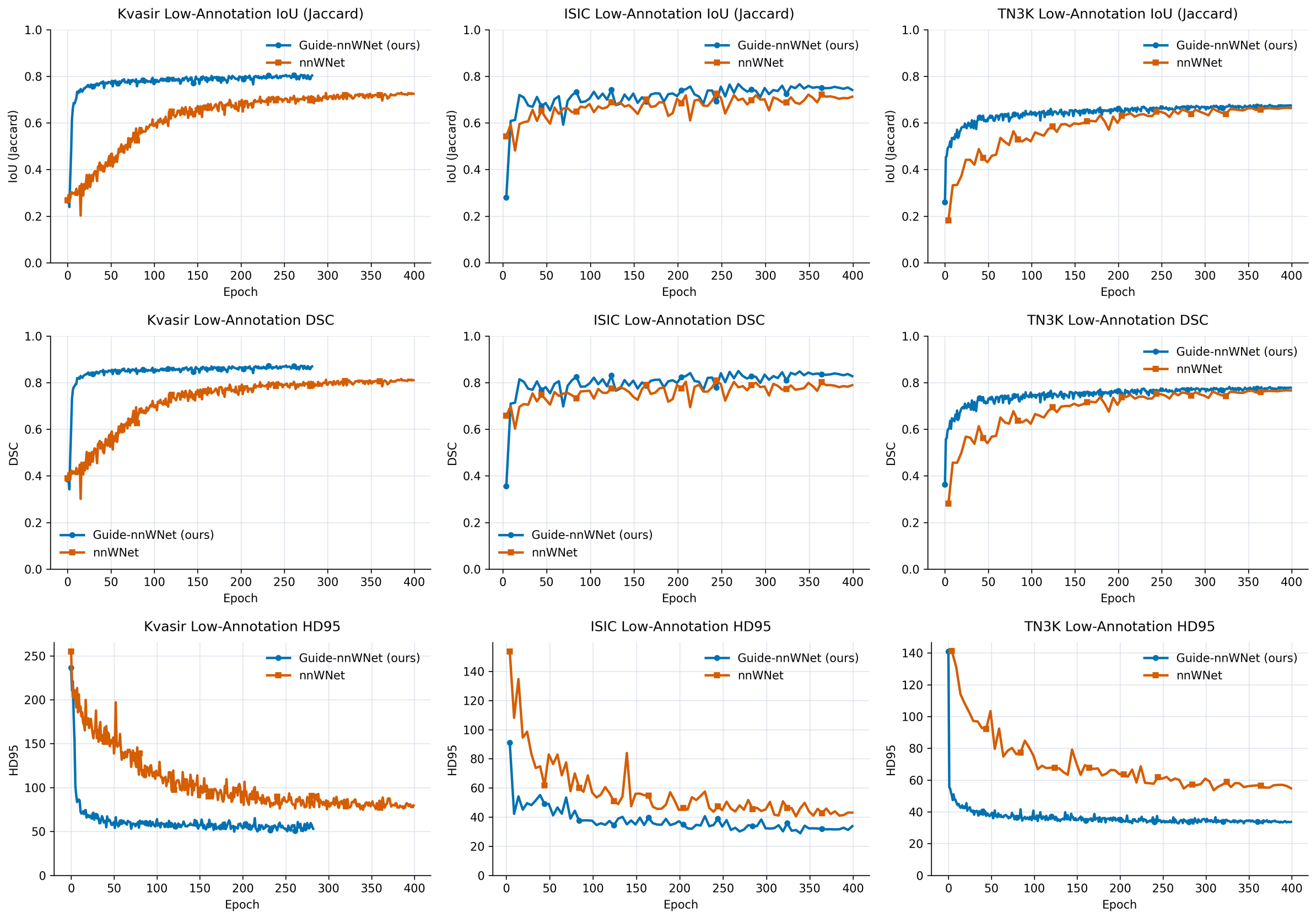}
    \captionsetup{hypcap=false}
    \captionof{figure}[Low-annotation learning curves with 30\% training masks.]{\textbf{Low-annotation learning curves with 30\% training masks.} Columns show Kvasir-SEG, ISIC 2017, and TN3K; rows show IoU, DSC, and HD95. LUMOS-W converges faster than nnWNet and maintains stronger boundary-distance behavior under limited dense supervision.}
    \label{fig:low_annotation_curves}
\end{center}
}]


\subsection{Curve Observations}
\label{app:lowann_curve_observations}

Figure~\ref{fig:low_annotation_curves} compares LUMOS-W and nnWNet validation trajectories under the low-annotation setting on Kvasir-SEG, ISIC 2017, and TN3K. We report IoU, DSC, and HD95, where lower HD95 is better. Across the three datasets, LUMOS-W adapts to limited labels faster and reaches stronger available curve endpoints than the nnWNet baseline.

\paragraph{Faster adaptation from limited labels.}
The clearest pattern is the early-epoch separation between the two methods. On Kvasir-SEG, both methods start from similar overlap at epoch 0, but LUMOS-W reaches 74.9 IoU and 83.2 DSC by epoch 19, while nnWNet remains at 33.7 IoU and 45.7 DSC. The same trend is visible on TN3K: at epoch 4, LUMOS-W already obtains 49.7 IoU and 60.3 DSC, compared with 18.2 IoU and 28.1 DSC for nnWNet. On ISIC 2017, nnWNet is stronger at the first validation point, but LUMOS-W overtakes it by epoch 9 and keeps a consistent advantage afterwards. These curves indicate that the guided model uses the small annotated subset more efficiently, rather than only improving after long training.

\paragraph{Overlap accuracy improves consistently.}
At the available validation-curve endpoints, LUMOS-W obtains higher IoU and DSC on all three datasets. On Kvasir-SEG, the endpoint IoU/DSC improves from 72.6/81.1 to 80.4/87.0, even though the available LUMOS-W history ends at epoch 282 rather than 399. On the corrected ISIC 2017 official-validation split, LUMOS-W improves the endpoint IoU/DSC from 71.3/78.9 to 74.2/82.8. On TN3K, the endpoint overlap gap is smaller, improving IoU/DSC from 66.4/76.7 to 67.5/77.8, but the guided method remains consistently above nnWNet over most of training.

\paragraph{Boundary-distance curves also improve.}
The HD95 curves show that LUMOS-W is not only improving region overlap but also reducing boundary outliers in these low-annotation validation histories. Endpoint HD95 is reduced from 79.64 to 53.04 on Kvasir-SEG, from 43.07 to 33.76 on ISIC 2017, and from 54.85 to 33.68 on TN3K. The best observed HD95 values show the same pattern: 49.82 versus 74.69 on Kvasir-SEG, 28.93 versus 40.50 on ISIC 2017, and 32.59 versus 53.59 on TN3K. This is important for medical segmentation, where small overlap gains can hide clinically relevant boundary errors.

\paragraph{Dataset-specific behavior.}
Kvasir-SEG shows the largest overlap separation and the fastest plateau, suggesting that the guide branch is especially effective when the reduced training subset still contains enough appearance diversity for the DINO-backed representation to generalize. ISIC 2017 shows a more moderate but cleaner advantage under the corrected official-validation protocol: LUMOS-W has better best and endpoint IoU, DSC, and HD95, without relying on the previously invalid partial-validation setting. TN3K is the closest dataset at the validation endpoint in IoU and DSC, but LUMOS-W has a large and persistent HD95 advantage, indicating better localization stability even when region overlap becomes similar.

\paragraph{Takeaway.}
The low-annotation curves support the main claim that LUMOS-W is more label-efficient than nnWNet. It reaches useful validation performance earlier, maintains higher endpoint IoU and DSC, and gives lower endpoint HD95 across the available histories. The result is strongest on Kvasir-SEG, remains clear under the corrected ISIC 2017 protocol, and is most visible as a boundary-quality improvement on TN3K.

\paragraph{Protocol note.}
The Kvasir-SEG LUMOS-W curve available locally ends at epoch 282, while the corresponding nnWNet curve runs to epoch 399. For TN3K, these plots use the available low-annotation histories; a corrected official-test TN3K low-annotation rerun was not available locally when these observations were drafted. \MainReducedAnnotationTable{} remains the source for the final reduced-annotation benchmark summary.

\section{Supplementary Ablations and Visualizations}
\label{sec:appendix_tables_figures}

\providecommand{\MainInjectionAblationTable}{Table~\ref{tab:injection_ablation}}
\setlength{\floatsep}{5pt plus 1pt minus 1pt}
\setlength{\textfloatsep}{6pt plus 1pt minus 2pt}
\setlength{\intextsep}{5pt plus 1pt minus 1pt}
\captionsetup[figure]{skip=2pt}
\captionsetup[table]{skip=2pt}
\newcommand{\apptablesquash}{\vspace{-0.65em}}
\newcommand{\appfigtopsquash}{\vspace{-0.45em}}
\newcommand{\appfigcapsquash}{\vspace{-0.35em}}
\newcommand{\appfigbottomsquash}{\vspace{-0.65em}}

\noindent In the following appendix tables, $s_G$ denotes the side length of the spatial guide mask, i.e., an $s_G\times s_G$ guide, and $K$ denotes the number of Reg Book prototypes. These local completed sweeps complement the main averaged result tables and should be read as sensitivity diagnostics. We report a single DSC column, together with IoU and HD95, to keep the metric naming consistent with the main text.

\paragraph{Guide-mask resolution.}
Tables~\ref{tab:appendix_tn3k_guide_size} and~\ref{tab:appendix_kvasir_guide_size} extend the guide-resolution study beyond the compact main-text ablation. TN3K is sensitive to $s_G$: larger guide masks improve boundary distance in this local sweep, but the overlap metrics saturate after moderate resolutions. Kvasir-SEG is less sensitive in DSC and IoU, while HD95 varies more strongly, suggesting that increasing guide resolution alone does not guarantee better contour stability. These observations support treating the guide as a coarse spatial emphasis signal rather than a high-resolution segmentation mask.

\begin{center}
\begin{minipage}{\linewidth}
\centering
\captionsetup{hypcap=false}
\captionof{table}{TN3K guide-mask side-length ablation. $s_G$ denotes the guide-mask side length.}
\label{tab:appendix_tn3k_guide_size}
\footnotesize
\setlength{\tabcolsep}{5pt}
\renewcommand{\arraystretch}{0.98}
\begin{tabular}{@{}lccc@{}}
\toprule
$s_G$ & DSC$\uparrow$ & IoU$\uparrow$ & HD95$\downarrow$ \\
\midrule
4 & 82.30 & 73.10 & 23.98 \\
8 & 82.70 & 73.30 & 24.65 \\
11 & 81.90 & 72.80 & 24.76 \\
16 & 83.10 & 74.00 & 24.22 \\
22 & 83.10 & 73.70 & 24.08 \\
32 & 82.20 & 72.80 & 25.15 \\
44 & 82.60 & 73.50 & 23.92 \\
64 & 82.90 & 73.70 & 23.43 \\
\bottomrule
\end{tabular}
\end{minipage}
\end{center}
\apptablesquash

\begin{center}
\begin{minipage}{\linewidth}
\centering
\captionsetup{hypcap=false}
\captionof{table}{Kvasir-SEG guide-mask side-length ablation. $s_G$ denotes the guide-mask side length.}
\label{tab:appendix_kvasir_guide_size}
\footnotesize
\setlength{\tabcolsep}{5pt}
\renewcommand{\arraystretch}{0.98}
\begin{tabular}{@{}lccc@{}}
\toprule
$s_G$ & DSC$\uparrow$ & IoU$\uparrow$ & HD95$\downarrow$ \\
\midrule
8 & 90.10 & 83.90 & 35.52 \\
16 & 90.40 & 84.20 & 38.12 \\
24 & 90.30 & 84.20 & 41.31 \\
32 & 90.40 & 84.20 & 39.13 \\
44 & 90.80 & 85.00 & 35.29 \\
64 & 90.80 & 84.90 & 38.93 \\
\bottomrule
\end{tabular}
\end{minipage}
\end{center}
\apptablesquash

\paragraph{Reg Book capacity.}
Tables~\ref{tab:appendix_tn3k_regbook_size} and~\ref{tab:appendix_kvasir_regbook_size} vary the Reg Book prototype count $K$ while keeping the guide-generation pipeline unchanged. The results show small but non-monotonic changes: increasing $K$ can stabilize the regularity response up to a point, but simply enlarging the Reg Book does not consistently improve all metrics. This behavior is consistent with the main formulation, where the Reg Book is a compact regularity memory and not a semantic class dictionary.

\begin{center}
\begin{minipage}{\linewidth}
\centering
\captionsetup{hypcap=false}
\captionof{table}{TN3K Reg Book prototype-count ablation. $K$ denotes the number of Reg Book prototypes.}
\label{tab:appendix_tn3k_regbook_size}
\footnotesize
\setlength{\tabcolsep}{5pt}
\renewcommand{\arraystretch}{0.98}
\begin{tabular}{@{}lccc@{}}
\toprule
$K$ & DSC$\uparrow$ & IoU$\uparrow$ & HD95$\downarrow$ \\
\midrule
64 & 81.90 & 72.70 & 25.41 \\
128 & 82.50 & 73.40 & 24.86 \\
256 & 82.50 & 73.30 & 24.44 \\
512 & 82.50 & 73.30 & 23.86 \\
1024 & 82.60 & 73.50 & 24.69 \\
2048 & 82.70 & 73.40 & 25.07 \\
4096 & 82.60 & 73.40 & 24.52 \\
\bottomrule
\end{tabular}
\end{minipage}
\end{center}
\apptablesquash

\begin{center}
\begin{minipage}{\linewidth}
\centering
\captionsetup{hypcap=false}
\captionof{table}{Kvasir-SEG Reg Book prototype-count ablation with available completed local runs. $K$ denotes the number of Reg Book prototypes.}
\label{tab:appendix_kvasir_regbook_size}
\footnotesize
\setlength{\tabcolsep}{5pt}
\renewcommand{\arraystretch}{0.98}
\begin{tabular}{@{}lccc@{}}
\toprule
$K$ & DSC$\uparrow$ & IoU$\uparrow$ & HD95$\downarrow$ \\
\midrule
64 & 90.50 & 84.30 & 39.94 \\
128 & 90.70 & 84.60 & 40.47 \\
256 & 90.00 & 83.80 & 39.22 \\
\bottomrule
\end{tabular}
\end{minipage}
\end{center}
\apptablesquash

\paragraph{Guidance application and model size.}
Table~\ref{tab:appendix_kvasir_mode} stresses how the guide is applied in an auxiliary completed Kvasir-SEG run. Hard masking collapses performance, indicating that the guide should not be treated as a binary constraint. Concatenation can be competitive on overlap in this auxiliary run, but it mixes guidance information into the feature stream and therefore weakens the role separation emphasized by LUMOS. The default soft-gating design is retained because the main matched ablation in \MainInjectionAblationTable{} supports soft gating over direct feature fusion, and because soft gating keeps the VFM-derived signal as spatial emphasis while leaving semantic prediction to the medical backbone. Table~\ref{tab:appendix_parameter_counts} reports parameter counts for transparency.

\begin{center}
\begin{minipage}{\linewidth}
\centering
\captionsetup{hypcap=false}
\captionof{table}{Kvasir-SEG guidance-application mode ablation.}
\label{tab:appendix_kvasir_mode}
\footnotesize
\setlength{\tabcolsep}{4pt}
\renewcommand{\arraystretch}{0.98}
\begin{tabular}{@{}lccc@{}}
\toprule
Mode & DSC$\uparrow$ & IoU$\uparrow$ & HD95$\downarrow$ \\
\midrule
Soft gate (ours) & 90.20 & 84.00 & 40.29 \\
Hard mask & 0.00 & 0.00 & 830.60 \\
Concatenate & 91.00 & 84.90 & 38.31 \\
\bottomrule
\end{tabular}
\end{minipage}
\end{center}
\apptablesquash

\begin{center}
\begin{minipage}{\linewidth}
\centering
\captionsetup{hypcap=false}
\captionof{table}{Parameter counts for implemented and external baseline methods. For frozen-encoder models, trainable parameters reflect the default fine-tuning policy used in our runs.}
\label{tab:appendix_parameter_counts}
\footnotesize
\setlength{\tabcolsep}{4pt}
\renewcommand{\arraystretch}{0.94}
\begin{tabular}{@{}lrr@{}}
\toprule
Method & Total (M) & Trainable (M) \\
\midrule
UNet & 7.76 & 7.76 \\
nnWNet & 7.04 & 7.04 \\
nnUNet-style & 71.82 & 71.82 \\
SwinUNet & 27.17 & 27.17 \\
SegDINO & 24.07 & 2.48 \\
LUMOS-U-DINO & 29.90 & 8.30 \\
LUMOS-W-DINO & 28.97 & 7.38 \\
DINO-Fusion nnWNet & 28.84 & 7.24 \\
LUMOS-U-SigLIP2 & 102.96 & 8.40 \\
LUMOS-W-SigLIP2 & 102.28 & 7.73 \\
H2Former & 33.68 & 33.68 \\
SAM2-UNet & 29.55 & 2.70 \\
\bottomrule
\end{tabular}
\end{minipage}
\end{center}
\apptablesquash

\begin{figure}[!t]
\centering
\appfigtopsquash
\includegraphics[width=0.96\linewidth]{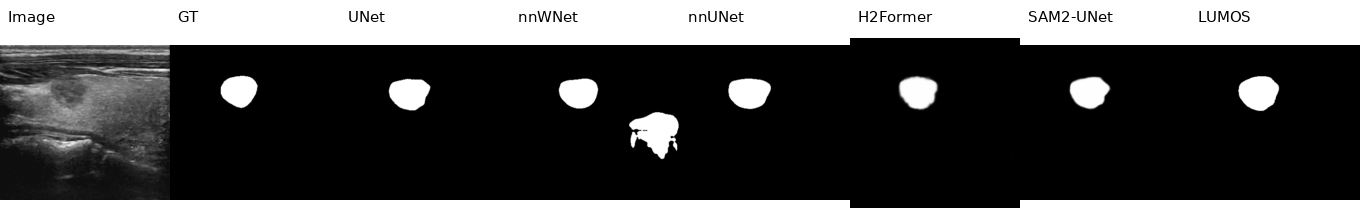}
\appfigcapsquash
\caption{TN3K held-out prediction comparison from local checkpoints. UNet, nnWNet, and LUMOS were regenerated from local checkpoints; LUMOS uses the completed guide-size run with $s_G=64$.}
\label{fig:appendix_tn3k_baseline_predictions}
\appfigbottomsquash
\end{figure}

\begin{figure}[t]
\centering
\appfigtopsquash
\includegraphics[width=0.88\linewidth]{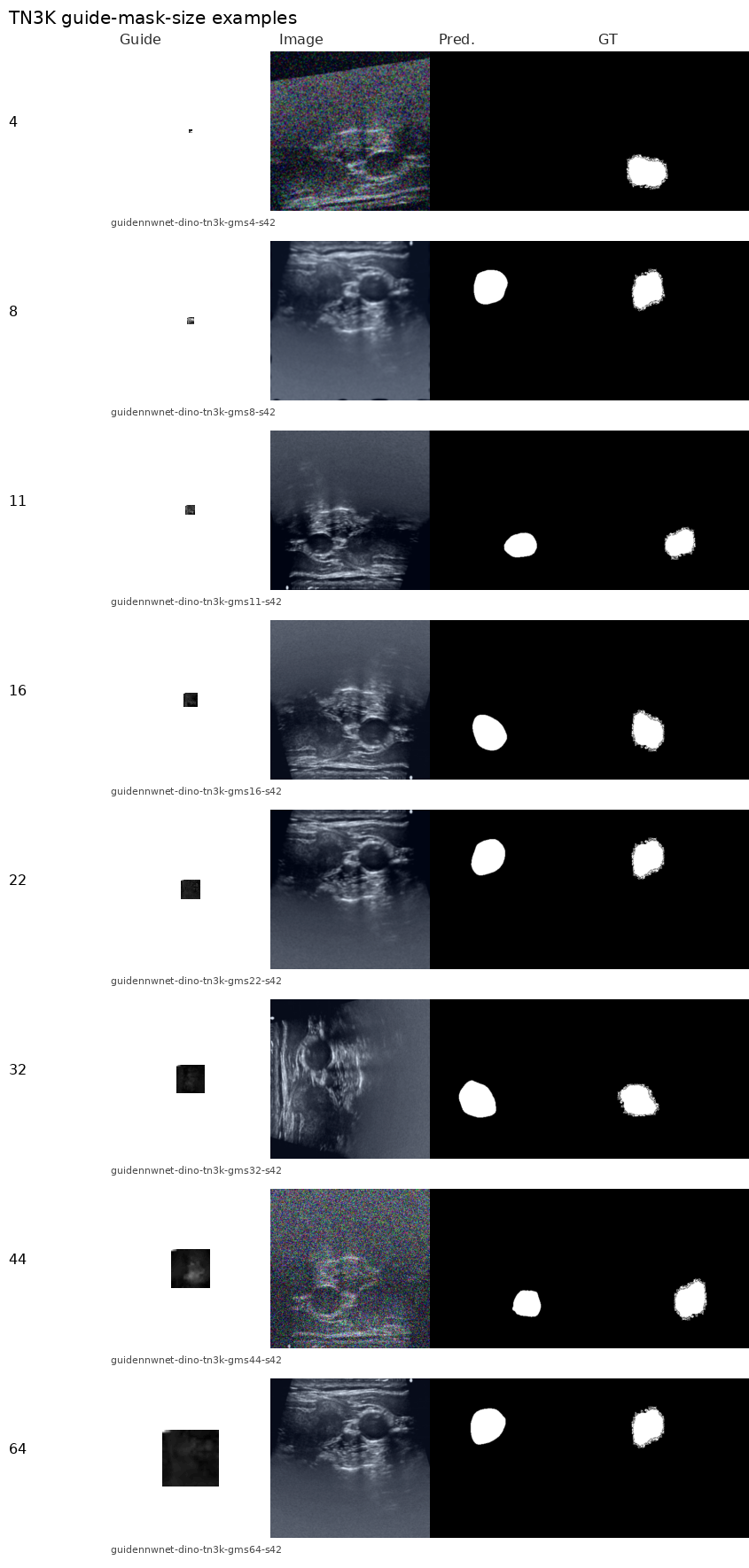}
\appfigcapsquash
\caption{TN3K training visualizations for guide-mask side-length settings $s_G$.}
\label{fig:appendix_tn3k_guide_size_examples}
\appfigbottomsquash
\end{figure}

\begin{figure}[t]
\centering
\appfigtopsquash
\includegraphics[width=0.90\linewidth]{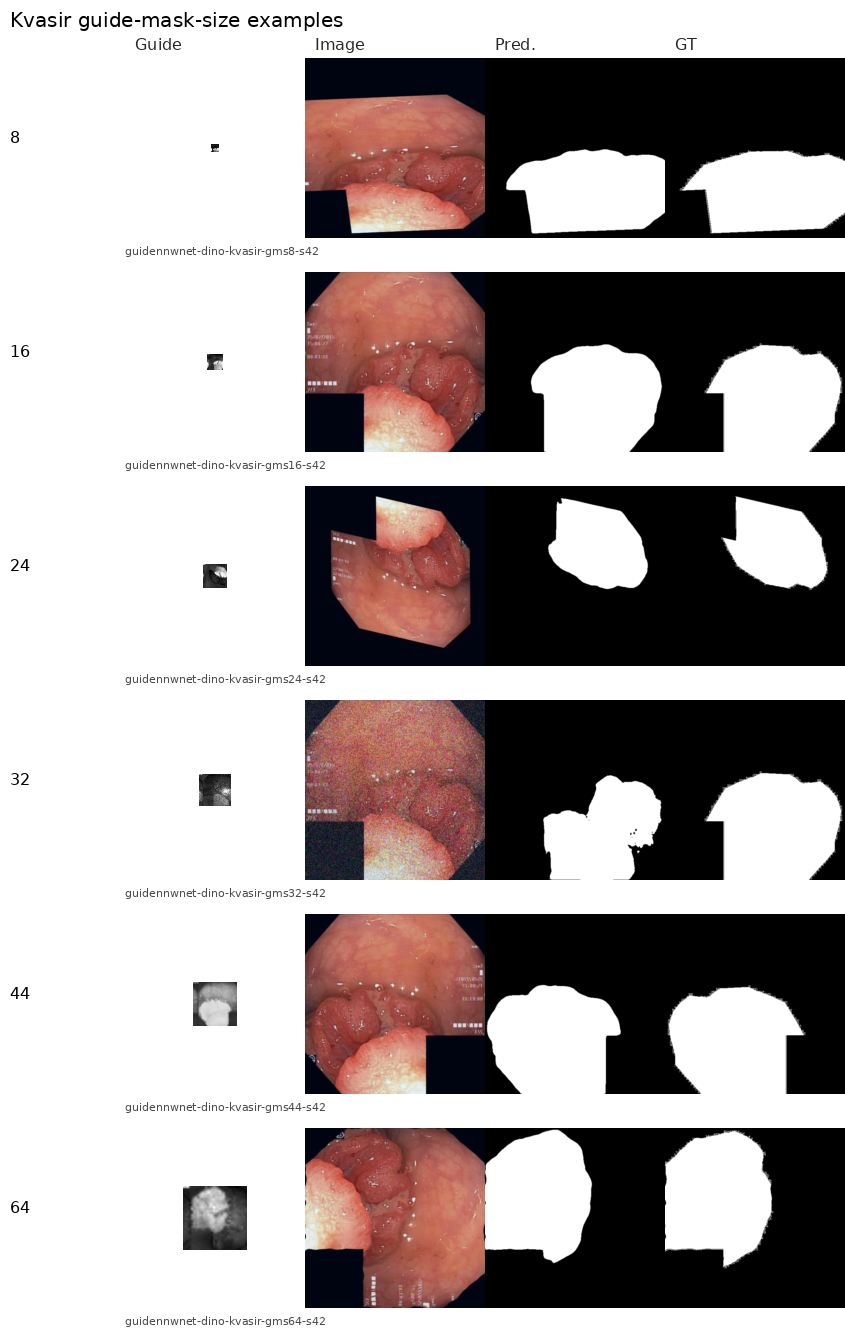}
\appfigcapsquash
\caption{Kvasir-SEG training visualizations for guide-mask side-length settings $s_G$.}
\label{fig:appendix_kvasir_guide_size_examples}
\appfigbottomsquash
\end{figure}

\begin{figure}[t]
\centering
\appfigtopsquash
\includegraphics[width=0.86\linewidth]{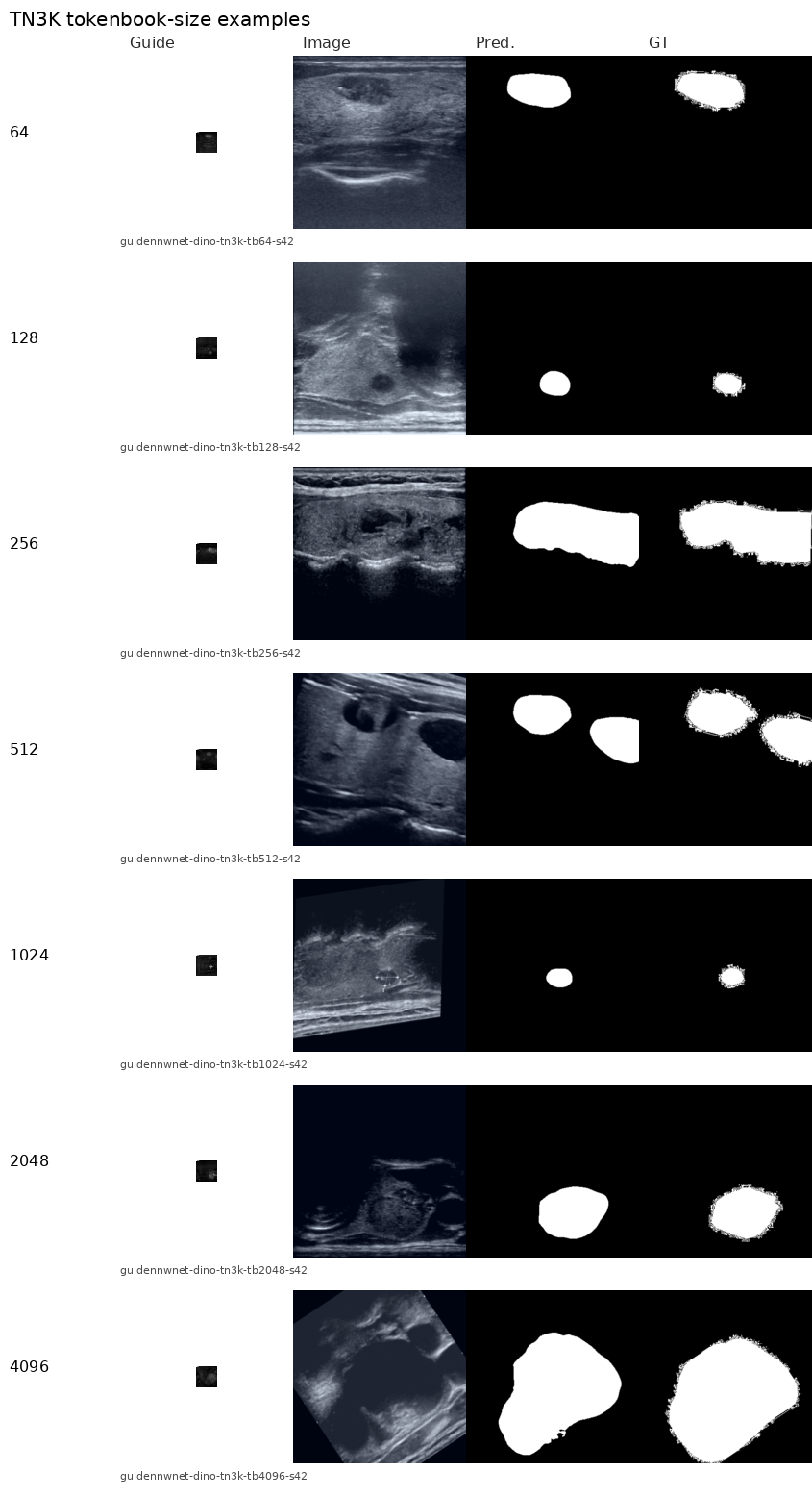}
\appfigcapsquash
\caption{TN3K training visualizations for Reg Book prototype-count settings $K$.}
\label{fig:appendix_tn3k_regbook_examples}
\appfigbottomsquash
\end{figure}

\begin{figure}[t]
\centering
\appfigtopsquash
\includegraphics[width=0.86\linewidth]{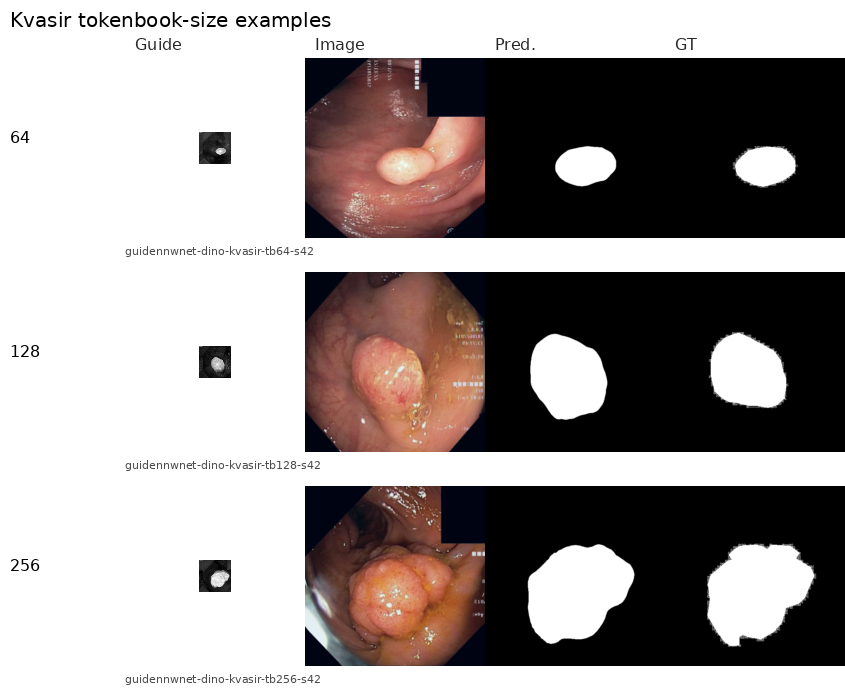}
\appfigcapsquash
\caption{Kvasir-SEG training visualizations for Reg Book prototype-count settings $K$ with available completed local runs.}
\label{fig:appendix_kvasir_regbook_examples}
\appfigbottomsquash
\end{figure}

\begin{figure}[t]
\centering
\appfigtopsquash
\includegraphics[width=0.86\linewidth]{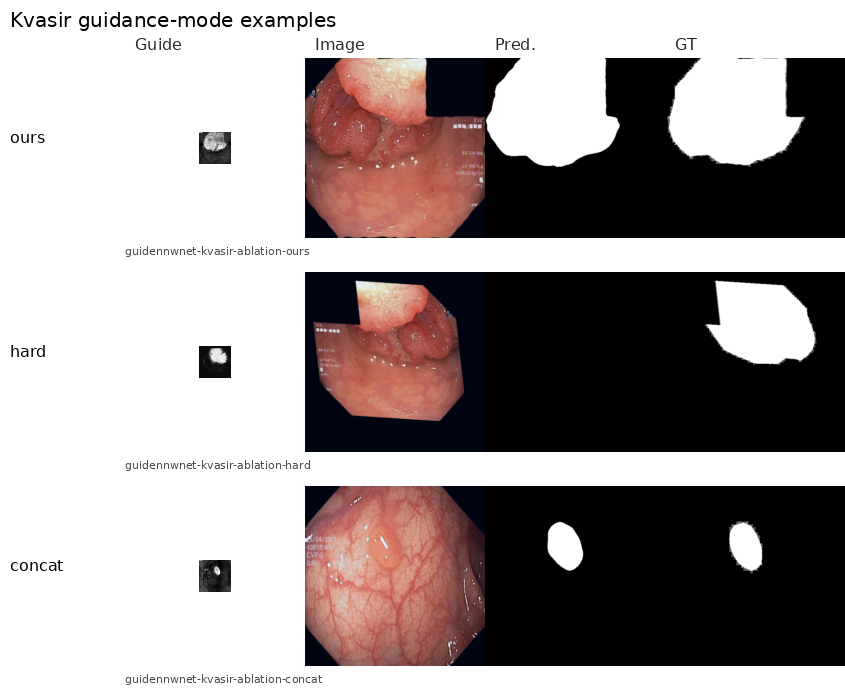}
\appfigcapsquash
\caption{Kvasir-SEG training visualizations for guidance-application modes.}
\label{fig:appendix_kvasir_mode_examples}
\appfigbottomsquash
\end{figure}

\paragraph{Qualitative observations.}
Figures~\ref{fig:appendix_tn3k_baseline_predictions}--\ref{fig:appendix_kvasir_mode_examples} provide supplementary qualitative examples for the ablations. The prediction comparison illustrates that LUMOS changes the target-region emphasis without using the guide as the final mask. The guide-size and Reg Book examples further show that the guide usually captures coarse target support rather than exact contours, which aligns with the HD95 sensitivity observed in the quantitative appendix tables.

\section{Additional Clarifications}
\label{sec:appendix_qa}

\paragraph{Q1. How is LUMOS different from using a foundation model as a medical segmentor?}
LUMOS does not use the frozen VFM as the final semantic predictor, decoder backbone, or post-processing module. The frozen VFM provides patch-level token responses, Pathfinder converts those responses into a soft spatial guide, and the medical segmentation backbone still produces the final mask. This separates spatial prior extraction from anatomical prediction.

\paragraph{Q2. Is the Reg Book just a prototype segmentation head?}
The Reg Book is prototype-like, but its role is deliberately restricted. Its prototypes represent reusable visual regularity patterns rather than anatomical classes, and their aggregated response forms a low-resolution guide mask. The guide is supervised during training, but it is not used as a final segmentation mask or a hard inference-time constraint.

\paragraph{Q3. Why use soft gating instead of direct feature fusion?}
Direct fusion forces natural-image VFM features into the medical backbone feature space, which may introduce semantic conflict. Inspiror instead transfers only spatial emphasis through a continuous guide mask, leaving anatomical classification and boundary refinement to the medical backbone. The Kvasir-SEG injection ablation supports this design: direct fusion degrades the nnWNet-based baseline, while soft gating improves IoU, Dice, and HD95.

\paragraph{Q4. Does guide supervision contradict the claim of using latent VFM priors?}
The guide is not claimed to be an unsupervised or zero-shot final segmentor. Training labels calibrate the Reg Book readout so that frozen VFM token responses become task-aware spatial guidance. At inference time, LUMOS uses the image, frozen VFM tokens, the learned Reg Book, and the medical backbone; no ground-truth mask is used, and the guide remains a soft modulation signal rather than a prediction to be evaluated directly.

\paragraph{Q5. Why are DINO and SigLIP not equally effective?}
The results indicate that the benefit of LUMOS depends on whether a VFM preserves patch-level pattern relevance for the target morphology. DINO provides stable matched-backbone Dice gains across the tested settings. SigLIP shows mixed behavior, which is consistent with its coarser token granularity and image-text representation objective producing less reliable local medical guidance in some datasets.

\paragraph{Q6. Why is LUMOS not always best on HD95?}
LUMOS is designed as a region-localization aid rather than a boundary-specialized decoder. Dice and IoU mainly measure region overlap, while HD95 is sensitive to isolated contour outliers. A low-resolution guide can improve target-region coverage while still leaving fine boundary accuracy to the backbone, decoder, and dataset-specific contour quality.

\paragraph{Q7. What does ``universal'' mean in this work?}
The term refers to transferable low-level visual regularities that can appear across token-based VFMs and medical image domains, such as contrast, texture change, boundary continuity, and region coherence. It should not be read as a claim that every VFM, dataset, metric, or segmentation setting will improve. The SigLIP and HD95 results explicitly bound the claim.

\paragraph{Q8. How should LUMOS be positioned relative to SAM-style medical foundation models?}
SAM-style systems are foundation-model segmentors or prompt/adaptation frameworks. LUMOS is complementary: it targets automatic segmentation with conventional medical backbones and uses frozen VFM tokens only to generate spatial guidance. The comparison therefore evaluates whether VFM-derived guidance can improve a medical segmentor, not whether LUMOS replaces promptable universal segmentation systems.

\paragraph{Q9. Is VFM-specific pattern enhancement a core contribution?}
No. SPE is treated as an auxiliary analysis. The main evidence for LUMOS comes from the frozen-VFM matched-backbone comparisons, where the VFM encoder is kept fixed and the Reg Book plus segmentation backbone are trained. SPE can improve some settings, but its mixed behavior supports the manuscript's caution that lightly shifting the VFM token space can also disturb useful frozen priors.

\paragraph{Q10. What is the current scope of evidence?}
The experiments cover three public 2D binary medical segmentation datasets: Kvasir-SEG, ISIC 2017, and TN3K. They support the claim that frozen VFM token spaces can provide useful spatial priors for these settings. Extension to 3D volumes, multi-class segmentation, external clinical cohorts, and prompt-based workflows remains outside the present evidence and should be treated as future work.

\end{document}